\documentclass[11pt,a4paper]{article}
\usepackage[hyperref]{emnlp-ijcnlp-2019}
\usepackage{times}
\usepackage{latexsym}
\usepackage{enumitem}
\usepackage{amsmath}
\usepackage{graphicx, multirow}
\usepackage{color}
\usepackage{amssymb}

\usepackage{url}
\newcommand{\ignore}[1]{}

\aclfinalcopy   

\title{Modeling Graph Structure in Transformer\\ for Better AMR-to-Text Generation}

\author{Jie Zhu$^1$\hspace{1.5cm}Junhui Li$^1$\thanks{Corresponding Author: Junhui Li.}\hspace{1.5cm}Muhua Zhu$^{2}$\\\textbf{Longhua Qian}$^1$\hspace{1cm}\textbf{Min Zhang}$^1$\hspace{1cm} \textbf{Guodong Zhou}$^1$\\
$^1$School of Computer Science and Technology, Soochow University, Suzhou, China\\
$^2$Alibaba Group, Hangzhou, China\\
{zhujie951121@gmail.com},
{\{lijunhui, qianlonghua, minzhang, gdzhou\}@suda.edu.cn}\\
{muhua.zmh@alibaba-inc.com}
}

\date{}

\begin{document}
\maketitle  

\begin{abstract}
Recent studies on AMR-to-text generation often formalize the task as a sequence-to-sequence (seq2seq) learning problem by converting an Abstract Meaning Representation (AMR) graph into a word sequence. Graph structures are further modeled into the seq2seq framework in order to utilize the structural information in the AMR graphs. However, previous approaches only consider the relations between directly connected concepts while ignoring the rich structure in AMR graphs. In this paper we eliminate such a strong limitation and propose a novel structure-aware self-attention approach to better modeling the relations between indirectly connected concepts in the  state-of-the-art seq2seq model, i.e., the \textit{Transformer}. In particular, a few different methods are explored to learn structural representations between two concepts. Experimental results on English AMR benchmark datasets show that our approach significantly outperforms the state of the art with 29.66 and 31.82 BLEU scores on LDC2015E86 and LDC2017T10, respectively. To the best of our knowledge, these are the best results achieved so far by supervised models on the benchmarks. 

\end{abstract}

\section{Introduction}
\begin{figure}[]
\begin{center}
\includegraphics[width=3.0in]{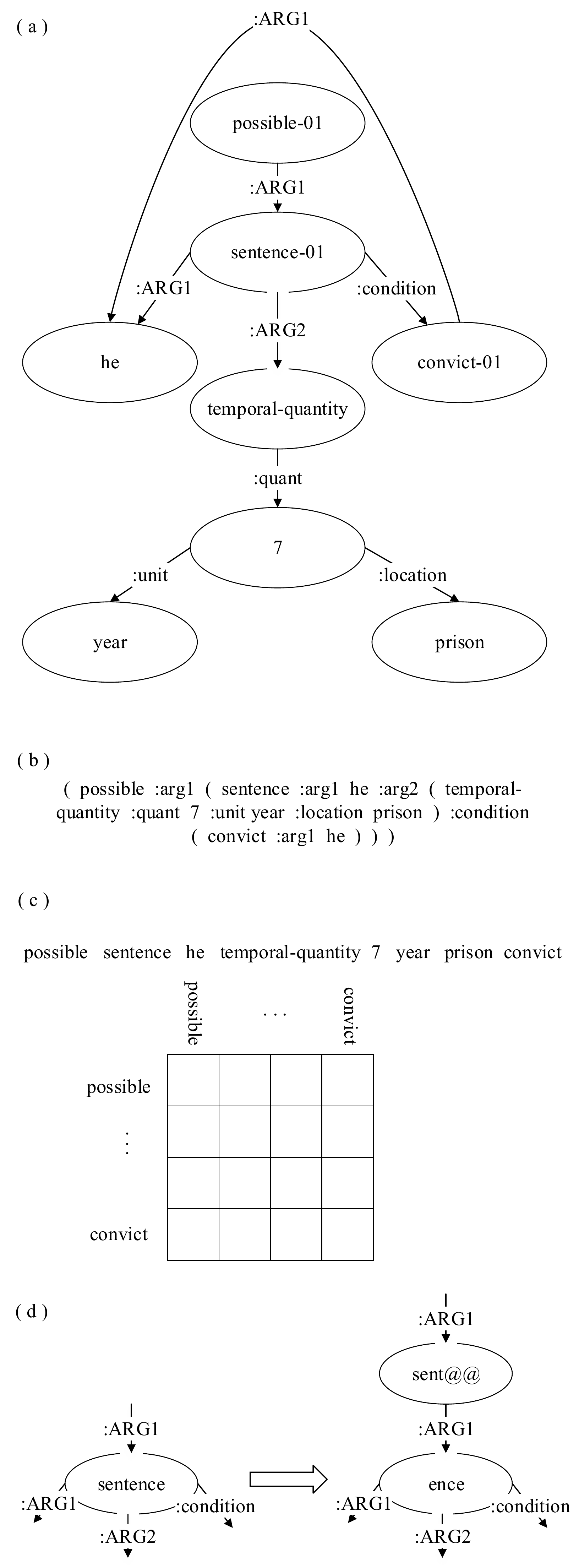}
\end{center}
\caption{(a) An example of AMR graph for the sentence of \textit{He could be sentenced to 7 years in prison if convicted.} (b) input to our baseline system, the seq2seq Transformer. (c) input to our proposed system based on structure-aware self-attention. (d) An example of graph structure extensions to sub-word units.} \label{fig:amr_example}
\end{figure}

AMR-to-text generation is a task of automatically generating a natural language sentence from an Abstract Meaning Representation (AMR) graph. Due to the importance of AMR as a widely adopted semantic formalism in representing the meaning of a sentence
~\cite{banarescu_etal:13}, AMR has become popular in semantic representation and AMR-to-text generation has been drawing more and more attention in the last decade. As the example in Figure~\ref{fig:amr_example}(a) shows, nodes, such as \textit{he} and \textit{convict-01}, represent semantic concepts and edges, such as ``:ARG1" and ``:quant", refer to semantic relations between the concepts. Since two concepts close in an AMR graph may map into two segments that are distant in the corresponding sentence, AMR-to-text generation is challenging. For example in Figure~\ref{fig:amr_example}, the neighboring concepts \textit{he} and \textit{convict-01} correspond to the words \textit{he} and \textit{convicted} which locate at the different ends of the sentence. 

To address the above mentioned challenge, recent studies on AMR-to-text generation regard the task as a sequence-to-sequence (seq2seq) learning problem by properly linearizing an AMR graph into a sequence~\cite{konstas_etal_acl:17}. Such an input representation, however, is apt to lose useful structural information due to the removal of reentrant structures for linearization. To better model graph structures, previous studies propose various graph-based seq2seq models to incorporate graphs as an additional input representation~\cite{song_etal_acl:18, beck_etal_acl:18, damonte_etal_naacl:19}. Although such graph-to-sequence models can achieve the state-of-the-art results, they focus on modeling one-hop relations only. That is, they only model concept pairs connected directly by an edge~\cite{song_etal_acl:18, beck_etal_acl:18}, and as a result, ignore explicit structural information of indirectly connected concepts in AMR graphs, e.g. the relation between concepts \textit{he} and \textit{possible} in Figure~\ref{fig:amr_example}. 

To make better use of structural information in an AMR graph, we attempt to model arbitrary concept pairs no matter whether directly connected or not. To this end, we extend the encoder in the state-of-the-art seq2seq model, i.e., the Transformer~\cite{vaswani_etal_nips:17} and propose structure-aware self-attention encoding approach. In particular, several distinct methods are proposed to learn structure representations for the new self-attention mechanism. Empirical studies on two English benchmarks show that our approach significantly advances the state of the art for AMR-to-text generation, with the performance improvement of 4.16 BLEU score on LDC2015E86 and 4.39 BLEU score on LDC2017T10 respectively over the strong baseline. Overall, this paper makes the following contributions.
\begin{itemize}
\item To the best of our knowledge, this is the first work that applies the Transformer to the task of AMR-to-text generation. On the basis of the Transformer, we build a strong baseline that reaches the state of the art.
\item We propose a new self-attention mechanism to incorporate richer structural information in AMR graphs. Experimental results on two benchmarks demonstrate the effectiveness of the proposed approach.  
\item Benefiting from the strong baseline and the structure-aware self-attention mechanism, we greatly advance the state of the art in the task.
\end{itemize}

\section{AMR-to-Text Generation with Graph Structure Modeling}
\label{sect:approach}
We start by describing the implementation of our baseline system, a state-of-the-art seq2seq model which is originally used for neural machine translation and syntactic parsing~\cite{vaswani_etal_nips:17}. Then we detail the proposed approach to incorporating structural information from AMR graphs. 
\subsection{Transformer-based Baseline}
\textbf{Transformer}: Our baseline system builds on the Transformer which employs an encoder-decoder framework, consisting of stacked encoder and decoder layers. Each encoder layer has two sublayers: self-attention layer followed by a position-wise feed forward layer. Self-attention layer employs multiple attention heads and the results from each attention head are concatenated and transformed to form the output of the self-attention layer. Each attention head uses scaled dot-product attention which takes a sequence $x=(x_1,\cdots,x_n)$ of $n$ elements as input 
and computes a new sequence $z=(z_1, \cdots, z_n)$ of the same length:
\begin{equation}
\label{equ:self-attention}
    z = Attention\left(x\right)
\end{equation}
where $x_i\in \mathbb{R}^{d_x}$  and $z\in\mathbb{R}^{n\times d_z}$. Each output element $z_i$ is a  weighted sum of a linear transformation of input elements:
\begin{equation}
\label{equ:z_i}
    z_i = \displaystyle\sum_{j=1}^{n} \alpha_{ij}\left(x_jW^V\right)
\end{equation}
where $W^V\in\mathbb{R}^{d_x\times d_z}$ is matrix of parameters. The vectors $\alpha_i=\left(\alpha_{i1},\cdots,\alpha_{in}\right)$ in Equation 2 are obtained by the self-attention model, which captures the correspondences between $x_i$ and others. Specifically, the attention weight $\alpha_{ij}$ of each element $x_j$ is computed using a softmax function:

\begin{equation}
    \alpha_{ij} = \frac{exp(e_{ij})}{\sum_{k=1}^{n}exp(e_{ik})}
\end{equation}
where
\begin{equation}
\label{equ:e_ij}
    e_{ij} = \frac{\left(x_iW^Q\right)\left(x_jW^K\right)^{T}}{\sqrt{d_z}}
\end{equation}
is an alignment function which measures how well the input elements $x_i$ and $x_j$ match. $W^Q, W^K\in\mathbb{R}^{d_x\times d_z}$ are parameters to be learned.

\noindent\textbf{Input Representation}: We use the depth-first traversal strategy as in Konstas et al.~\shortcite{konstas_etal_acl:17} to linearize AMR graphs and to obtain simplified AMRs. We remove variables, wiki links and sense tags before linearization. Figure~\ref{fig:amr_example}(b) shows an example linearization result for the AMR graph in Figure~\ref{fig:amr_example}(a). Note that the reentrant concept \textit{he} in Figure~\ref{fig:amr_example} (a) maps to two different tokens in the linearized sequence.

\noindent\textbf{Vocabulary}: Training AMR-to-text generation systems solely on labeled data may suffer from data sparseness. To attack this problem, previous works adopt techniques like anonymization to remove named entities and rare words~\cite{konstas_etal_acl:17}, or apply a copy mechanism~\cite{gulcehre_etal_acl:16} such that the models can learn to copy rare words from the input sequence. In this paper we instead use two simple yet effective techniques. One is to apply Byte Pair Encoding (BPE)~\cite{sennrich_etal_acl:16} to split words into smaller, more frequent sub-word units. The other is to use a shared vocabulary for both source and target sides. Experiments in Section ~\ref{sect:result} demonstrate the necessity of the techniques in building a strong baseline. 

\subsection{Modeling Graph Structures in Transformer}
\noindent\textbf{Input Representation:} We also use the depth-first traversal strategy to linearize AMR graphs and to obtain simplified AMRs which only consist of concepts. As shown in Figure~\ref{fig:amr_example} (c), the input sequence is much shorter than the input sequence in the baseline. Besides, we also obtain a matrix which records the graph structure between every concept pair, which implies their semantic relationship (Section\ref{sect:structure_representation}). 

\noindent\textbf{Vocabulary}: To be compatible with sub-words, we extend the original AMR graph, if necessary, to include the structures of sub-words. As \textit{sentence-01} in Figure~\ref{fig:amr_example}(a) is segmented into \textit{sent@@ ence-01}, we split the original node into two connected ones with an edge labeled as the incoming edge of the first unit. Figure~\ref{fig:amr_example}(d) shows the graph structure for sub-words \textit{sent@@ ence-01}.

\noindent\textbf{Structure-Aware Self-Attention}: Motivated by~\citet{shaw_etal_naacl:18}, we extend the conventional self-attention architecture to explicitly encode the relation between an element pair ($x_i, x_j$) in the alignment model by replacing Equation~\ref{equ:e_ij} with Equation~\ref{equ:new_e_ij}. Note that the relation $r_{ij}\in\mathbb{R}^{d_z}$ is the vector representation for element pair ($x_i, x_j$), and will be learned in Section~\ref{sect:structure_representation}. 
\begin{equation}
\label{equ:new_e_ij}
    e_{ij} = \frac{\left(x_iW^Q\right)\left(x_jW^K + r_{ij}W^{R}\right)^{T}}{\sqrt{d_z}}
\end{equation}
where $W^{R}\in\mathbb{R}^{d_z\times d_z}$ is a parameter matrix. Then, we update Equation~\ref{equ:z_i} accordingly to propagate structure information to the sublayer output by:

\begin{equation}
\label{equ:new_z_i}
    z_i = \displaystyle\sum_{j=1}^{n} \alpha_{ij}\left(x_jW^V + r_{ij}W^F\right)
\end{equation}
where $W^{F}\in\mathbb{R}^{d_z\times d_z}$ is a parameter matrix. 
\ignore{
Note that to enable an efficient implementation, we follow \citet{shaw_etal_naacl:18} and use simple addition to incorporate structure representations in Equqation~\ref{equ:new_e_ij} and Equation~\ref{equ:new_z_i}.
}

\subsection{Learning Graph Structure Representation for Concept Pairs}
\label{sect:structure_representation}
The above structure-aware self-attention is capable of incorporating graph structure between concept pairs. In this section, we explore a few methods to learn the representation for concept pairs. We use a sequence of edge labels, along the path from $x_i$ to $x_j$ to indicate the AMR graph structure between concepts $x_i$ and $x_j$.\footnote{While there may exist two or more paths connecting $x_i$ and $x_j$, we simply choose the shortest one.} In order to distinguish the edge direction, we add a direction symbol to each label with $\uparrow$ for climbing up along the path, and $\downarrow$ for going down. Specifically, for the special case of $i==j$, we use \textit{None} as the path. Table~\ref{tbl:path} demonstrates structural label sequences between a few concept pairs in Figure~\ref{fig:amr_example}. 

Now, given a structural path with a label sequence $s = s_1,\cdots,s_k$ and its $d_x$-sized corresponding label embedding sequence $l = l_1, \cdots, l_k$, we use the following methods to obtain its representation vector $r$, which maps to $r_{ij}$ in Equation~\ref{equ:new_e_ij} and Equation~\ref{equ:new_z_i}.

\begin{table}[]
    \centering
    \begin{tabular}{ll|r}
    \hline
         \bf $x_i$ & \bf $x_j$ & \bf Structural label sequence\\
    \hline
        he  & convict-01 & :ARG1$\uparrow$ \\ 
        he  & 7 & :ARG1$\uparrow$~ :ARG2$\downarrow$~ :quant$\downarrow$ \\
        he  & he & None \\
    \hline
    \end{tabular}
    \caption{Examples of structural path between a few concept pairs in Figure~\ref{fig:amr_example}.}
    \label{tbl:path}
\end{table}

\noindent\textbf{Feature-based} 

\noindent A natural way to represent the structural path is to view it as a string feature. To this end, we combine the labels in the structural path into a string. Unsurprisingly, this will end up with a large number of features. We keep the most frequent ones (i.e., 20K in our experiments) in the feature vocabulary and map all others into a special feature \textit{UNK}. Each feature in the vocabulary will be mapped into a randomly initialized vector.

\noindent\textbf{Avg-based} 

\noindent To overcome the data sparsity in the above feature-based method, we view the structural path as a label sequence. Then we simply use the averaged label embedding as the representation vector of the sequence, i.e., 

\begin{equation}
    r = \frac{\sum_{i=1}^{k}l_i}{k}
\end{equation}

\noindent\textbf{Sum-based} 

\noindent Sum-based method simply returns the sum of all label embeddings in the sequence, i.e., 

\begin{equation}
    r = \sum_{i=1}^{k}l_i
\end{equation}

\noindent\textbf{Self-Attention-based (SA-based for short)} 

\begin{figure}[t]
\begin{center}
\includegraphics[width=2.8in]{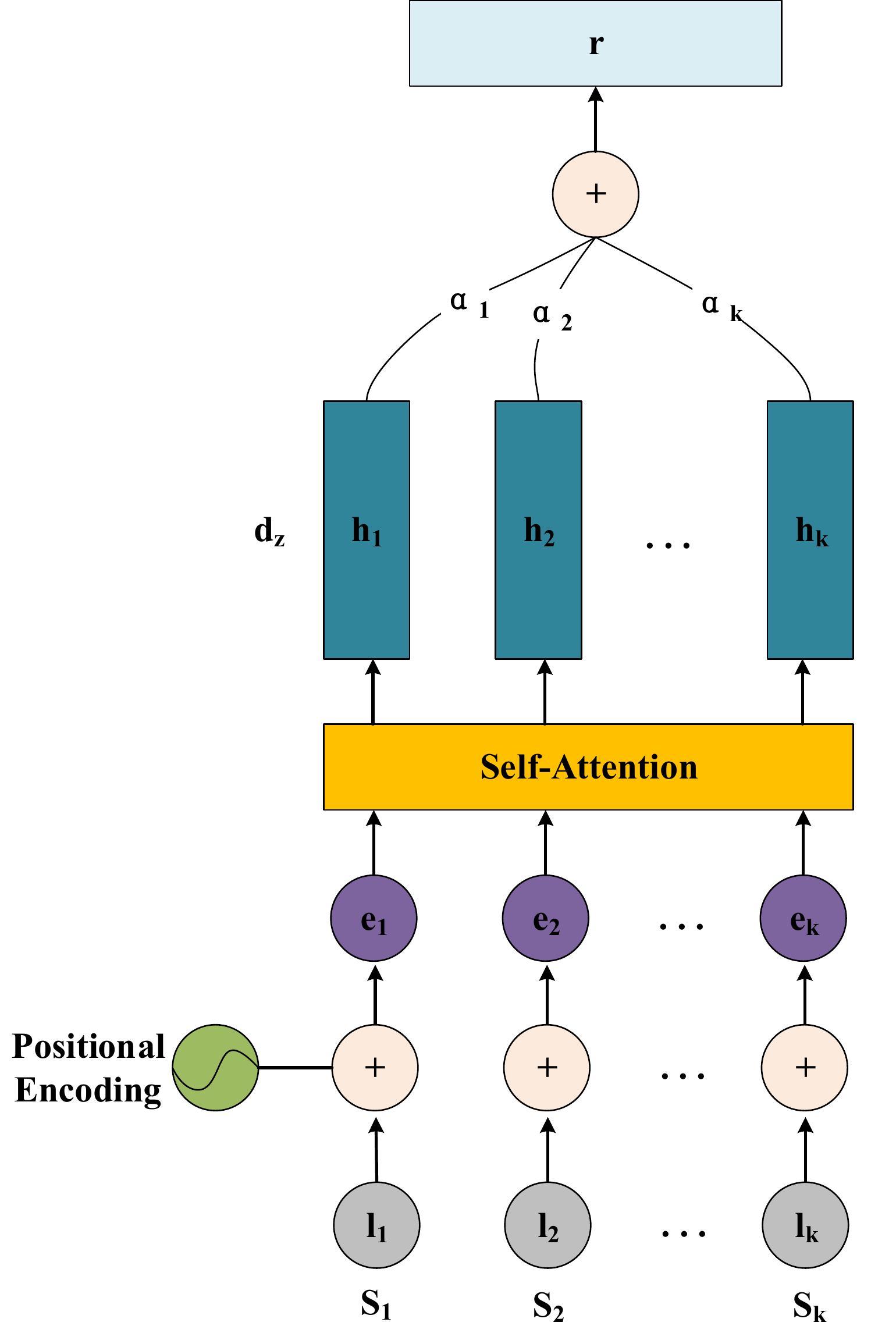}
\end{center}
\caption{Self-Attention-based method.} \label{fig:sa}
\end{figure}

\noindent As shown in Figure~\ref{fig:sa}, given the label sequence $s = s_1,\cdots,s_k$, we first obtain the sequence $e$, whose element is the addition of a word embedding and the corresponding position embedding. Then we use the self-attention, as presented in Eq.~\ref{equ:self-attention} to obtain its hidden states $h$, i.e,  $h=Attention(e)$\ignore{ $h = h_1,\cdots,h_k$}, where $h_i\in\mathbb{R}^{d_z}$. Our aim is to encode a variable length sentence into a $d_z$-sized vector. Motivated by \cite{lin_etal_iclr:17}, we achieve this by
choosing a linear combination of the $k$ vectors in $h$. Computing the linear combination requires an attention mechanism which takes the whole hidden states $h$ as input, and outputs a vector of weights $\alpha$:

\begin{equation}
\alpha = softmax(W^2\tanh(W^1h^T))    
\end{equation}
where $W^1\in\mathbb{R}^{d_w\times d_z}$ and $W^2\in\mathbb{R}^{d_w}$. Then the label sequence representation vector is the weighted sum of its hidden states:
\begin{equation}
    r = \sum_{i=1}^{k}\alpha_ih_i
\end{equation}

\noindent\textbf{CNN-based} 

\noindent Motivated by \cite{kalchbrenner_etal_acl:14}, we use convolutional neural network (CNN) to convolute the label sequence $l$ into a vector $r$, as follow:

\begin{equation}
\begin{split}
conv = Conv1D(kernel\_size &= (m),\\
strides &= 1,\\
filters &= d_z,\\
input\_shape &= d_z\\
activation &= 'relu')
\end{split}
\end{equation}
\begin{equation}
    r = conv\left(l\right)
\end{equation}
where kernel size $m$ is set to 4 in our experiments.

\section{Experimentation}
\subsection{Experimental Settings}
For evaluation of our approach, we use the sentences annotated with AMRs from the LDC release LDC2015E86 and LDC2017T10. The two datasets contain 16,833 and 36,521 training AMRs, respectively, and share 1,368 development AMRs and 1,371 testing AMRs. We segment words into sub-word units by BPE~\cite{sennrich_etal_acl:16} with 10K operations on LDC2015E86 and 20K operations on LDC2017T10. 

For efficiently learning graph structure representation for concept pairs (except the feature-based method), we limit the maximum label sequence length to 4 and ignore the labels exceeding the maximum. In SA-based method, we set the filter size $d_w$ as 128.

We use \textit{OpenNMT}~\cite{klein_etal:17} as the implementation of the Transformer seq2seq model.\footnote{\url{https://github.com/OpenNMT/OpenNMT-py}} In parameter setting, we set the number of layers in both the encoder and decoder to 6. For optimization we use Adam with $\beta1$ = 0.1~\cite{kingma_ba_iclr:15}. The number of heads is set to 8. In addition, we set the embedding and the hidden sizes to 512 and the batch token-size to 4096. Accordingly, the $d_x$ and $d_z$ in Section~\ref{sect:approach} are 64. In all experiments, we train the models for 300K steps on a single K40 GPU.

For performance evaluation, we use BLEU~\cite{papineni_etal_acl:02}, Meteor~\cite{banerjee_and_lavie:05,denkowski_and_lavie:14}, and CHRF++~\cite{popovic_wmt:17} as metrics. We report results of single models that are tuned on the development set.

We make our code available at \url{https://github.com/Amazing-J/structural-transformer}.


\subsection{Experimental Results}
\label{sect:result}

\begin{table}[t]
    \centering
    \begin{tabular}{l|ccc}
    \hline
    \bf Model & \bf BLEU & \bf Meteor & \bf CHRF++\\
    \hline
    Baseline &  24.93 & 33.20 & 60.30\\
    \hline
    -BPE  & 23.02 & 31.60 & 58.09  \\
    -Share Vocab.     & 23.24 & 31.78 & 58.43\\
    -Both & 18.77 & 28.04 & 51.88\\
    \hline
    \end{tabular}
    \caption{Ablation results of our baseline system on the LDC2015E86 development set.}
    \label{tbl:baseline}
\end{table}

We first show the performance of our baseline system. As mentioned before, BPE and sharing vocabulary are two techniques we applied to relieving data sparsity. Table~\ref{tbl:baseline} presents the results of the ablation test on the development set of LDC2015E86 by either removing BPE, or vocabulary sharing, or both of them from the baseline system. From the results we can see that BPE and vocabulary sharing are critical to building our baseline system (an improvement from 18.77 to 24.93 in BLEU), revealing the fact that they are two effective ways to address the issue of data sparseness for AMR-to-text generation. 
\begin{table*}
    \setlength{\tabcolsep}{3pt}
    \centering
    \begin{tabular}{c|c|ccc|c|ccc}
         \hline
         \multicolumn{2}{c|}{\multirow{2}{*}{\bf System}}& \multicolumn{4}{c|}{\bf LDC2015E86} & \multicolumn{3}{c}{\bf LDC2017T10}\\
         \cline{3-9}
         \multicolumn{2}{c|}{} & BLEU & Meteor & CHRF++ & \#P (M)& BLEU & Meteor & CHRF++ \\
         \hline
         \multicolumn{2}{c|}{Baseline} & 25.50 & 33.16 & 59.88 & 49.1 & 27.43 & 34.62 & 61.85\\
         \hline
         \multirow{5}{*}{\bf Our Approach} & feature-based & 27.23 & 34.53 & 61.55 & 49.4 & 30.18 & 35.83 & 63.20\\
         \cline{2-9}
         & avg-based & 28.37 & 35.10 & 62.29 & 49.1 & 29.56 & 35.24 & 62.86\\
         \cline{2-9}
         & sum-based & 28.69 & 34.97 & 62.05 & 49.1 & 29.92 & 35.68 & 63.04\\
         \cline{2-9}
         & SA-based & \textbf{29.66} & \textbf{35.45} & \textbf{63.00} & 49.3 & 31.54 & 36.02 & 63.84\\
         \cline{2-9}
         & CNN-based & 29.10 & 35.00 & 62.10 & 49.2 & \textbf{31.82} & \textbf{36.38} & \textbf{64.05}\\
         \hline
         \hline
         \multicolumn{8}{l}{Previous works with single models}\\
         \hline
         \multicolumn{2}{c|}{~\citet{konstas_etal_acl:17}$^\ast$} & 22.00 & - & - & - & - & - & -\\
         \multicolumn{2}{c|}{~\citet{cao_etal_naacl:19}$^\ast$} & 23.5 & - & - & - & 26.8 & - & -\\
         \multicolumn{2}{c|}{~\citet{song_etal_acl:18}$^\dagger$} & 23.30 & - & - & - & - & - & - \\
         \multicolumn{2}{c|}{~\citet{beck_etal_acl:18}$^\dagger$} & - & - & - & - & 23.3 & - & 50.4  \\
         \multicolumn{2}{c|}{~\citet{damonte_etal_naacl:19}$^\dagger$} & 24.40 & 23.60 & - & - & 24.54 & 24.07 & - \\
         \multicolumn{2}{c|}{~\citet{guo_etal_tacl:19}$^\dagger$} &  25.7 & - & - & - & 27.6 & - &  57.3 \\
         \multicolumn{2}{c|}{~\citet{song_etal_emnlp:16}$^\ddag$} & 22.44 & - & - & - & - & - & -\\
         \hline
         \multicolumn{8}{l}{Previous works with either ensemble models or unlabelled data, or both}\\
         \hline
         \multicolumn{2}{c|}{~\citet{konstas_etal_acl:17}$^\ast$} & 33.8 & - & - & - & - & - & -\\
         \multicolumn{2}{c|}{~\citet{song_etal_acl:18}$^\dagger$} & 33.0 & - & - & - & - & - & -\\
         \multicolumn{2}{c|}{~\citet{beck_etal_acl:18}$^\dagger$} & - & - & - & - & 27.5 & - & 53.5 \\
         \multicolumn{2}{c|}{~\citet{guo_etal_tacl:19}$^\dagger$} &  35.3 & - & - & - & -& - &  - \\
         \hline
    \end{tabular}
    \caption{Comparison results of our approaches and related studies on the test sets of LDC2015E86 and LDC2017T10. \#P indicates the size of parameters in millions. $^\ast$ indicates seq2seq-based systems while $^\dagger$ for graph-based models, and $^\ddag$ for other models. All our proposed systems are significant over the baseline at 0.01, tested by bootstrap resampling~\cite{koehn_emnlp:04}.}
    \label{tbl:amr_performance}
\end{table*}
\\
\indent Table~\ref{tbl:amr_performance} presents the comparison of our approach and related works on the test sets of LDC2015E86 and LDC2017T10. From the results we can see that the Transformer-based baseline outperforms most of graph-to-sequence models and is comparable with the latest work by ~\citet{guo_etal_tacl:19}. The strong performance of the baseline is attributed to the capability of the Transformer to encode global and implicit structural information in AMR graphs. By comparing the five methods of learning graph structure representations, we have the following observations.
\begin{itemize}
\item All of them achieve significant improvements over the baseline: the biggest improvements are 4.16 and 4.39 BLEU scores on LDC2015E86 and LDC2017T10, respectively.
\item Methods using continuous representations (such as SA-based and CNN-based) outperform the methods using discrete representations (such as feature-based).
\item Compared to the baseline, the methods have very limited affect on the sizes of model parameters (see the column of \textit{\#P (M)} in Table~\ref{tbl:amr_performance}).
\end{itemize}
 Finally, our best-performing models are the best among all the single and supervised models.

\section{Analysis}
In this section, we use LDC2017T10 as our benchmark dataset to demonstrate how our proposed approach achieves higher performance than the baseline. As representative, we use CNN-based method to obtain structural representation.

\subsection{Effect of Modeling Structural Information of Indirectly Connected Concept Pairs}

Our approach is capable of modeling arbitrary concept pairs no matter whether directly connected or not. To investigate the effect of modeling structural information of indirectly connected concept pairs, we ignore their structural information by mapping all structural label sequences between two indirectly connected concept pairs into \textit{None}. In this way, the structural label sequence for \textit{he} and \textit{7} in Table~\ref{tbl:path}, for example, will be \textit{None}.

\begin{table}[t]
    \centering
    \begin{tabular}{c|c} 
     \hline
     \bf System & \bf BLEU \\ 
     \hline
     Baseline  &  27.43 \\
     \hline
     Our approach  &  31.82 \\
     \hline
     No indirectly connected concept pairs & 29.92 \\ 
     \hline
    \end{tabular}
    \caption{Performance on the test set of our approach with or without modeling structural information of indirectly connected concept pairs.}
    \label{tbl:amr_indirect}
\end{table}

Table~\ref{tbl:amr_indirect} compares the performance of our approach with or without modeling structural information of indirectly connected concept pairs. It shows that by modeling structural information of indirectly connected concept pairs, our approach improves the performance on the test set from 29.92 to 31.82 in BLEU scores. It also shows that even without modeling structural information of indirectly connected concept pairs, our approach achieves better performance than the baseline.

\ignore{The proposed method of representing every concept pair relation relies on the explicit direct relations to compute the indirect relations, which is similar to what is done implicitly in any graph-based encoder. However, the graph-based encoder only considers modeling the directly associated concept pairs without modeling the indirect concept pairs. In order to reflect the necessity and impact of modeling indirect relationships, we removed the relationships between non-adjacent nodes and evaluated their performance. Table~\ref{tbl:amr_indirect} shows the results which indicate that modeling indirect relationship significantly outperforms not modeling indirect relationships.}

\ignore{
\begin{table}[h]
    \centering
    \begin{tabular}{ |c|c|c| } 
     \hline
           & LDC2015E86 & LDC2017T10 \\ 
     \hline
     FULL  & 29.66 & 31.82 \\
     \hline
     no-indirect & 26.28 & 29.92 \\ 
     \hline
    \end{tabular}
    \caption{Performance (in BLEU) on the test set.}
    \label{tab:amr_indirect}
\end{table}
}

\subsection{Effect on AMR Graphs with Different Sizes of Reentrancies}
\begin{figure}[t]
\begin{center}
\includegraphics[width=2.8in]{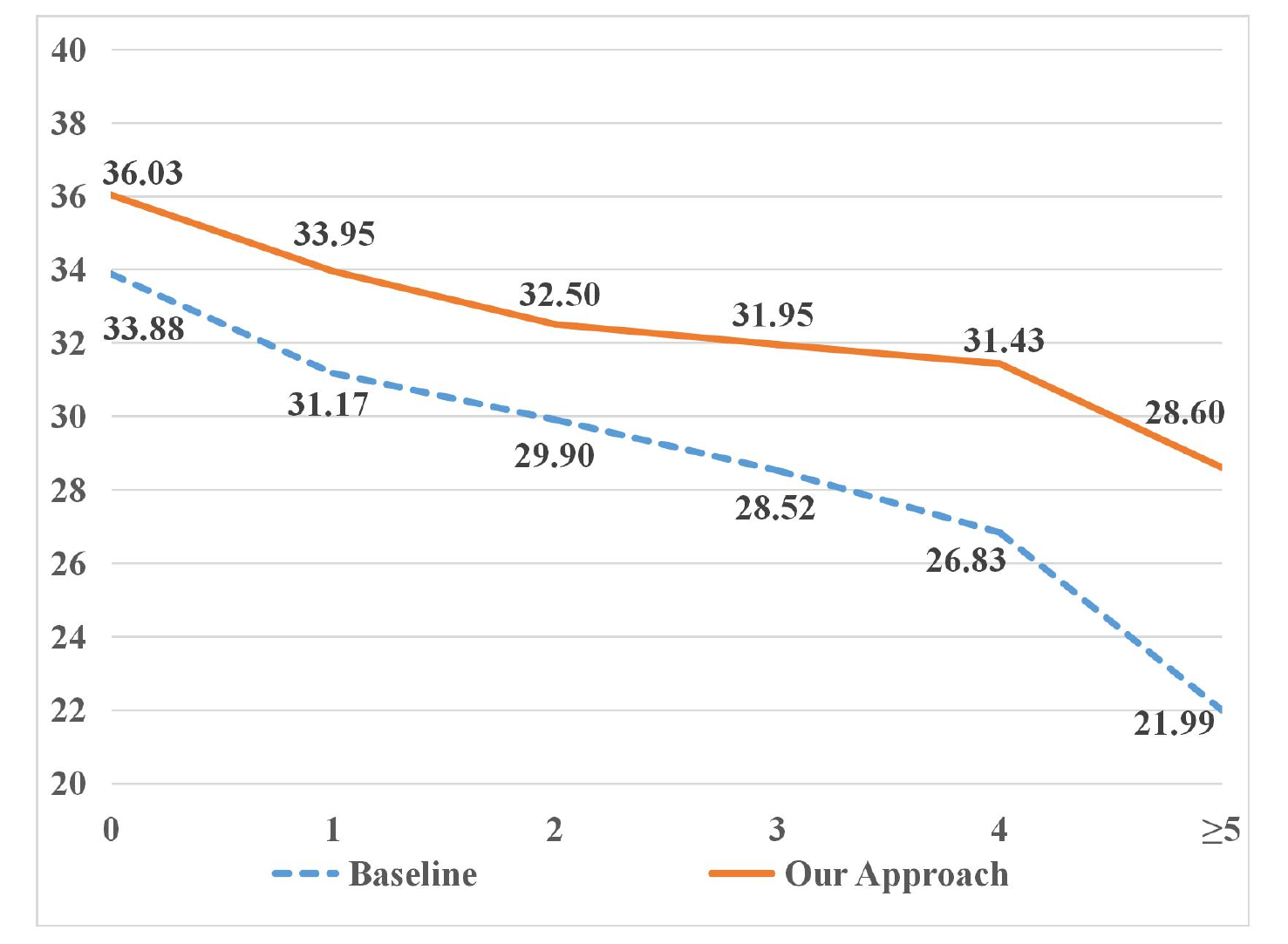}
\end{center}
\caption{Performance (in BLEU) on the test set with respect to the reentrancy numbers of the input AMR graphs.} \label{fig:reen}
\end{figure}

Linearizing an AMR graph into a sequence unavoidably loses information about reentrancies  (nodes with multiple parents). This poses a challenge for the baseline since there exists on obvious sign that the first \textit{he} and the second \textit{he}, as shown in Figure~\ref{fig:amr_example} (b), refer to the same person. By contrast, our approach models reentrancies explicitly. Therefore, it is expected that the benefit of our approach is more evident for those AMR graphs containing more reentrancies. To test this hypothesis, we partition source AMR graphs to different groups by their numbers of reentrancies and evaluate their performance respectively. As shown in Figure~\ref{fig:reen}, the performance gap between our approach and the baseline goes widest for AMR graphs with more than 5 reentrancies, on which our approach outperforms the baseline by 6.61 BLEU scores. 

\subsection{Effect on AMR Graphs with Different Sizes}
\begin{figure}[t]
\begin{center}
\includegraphics[width=2.8in]{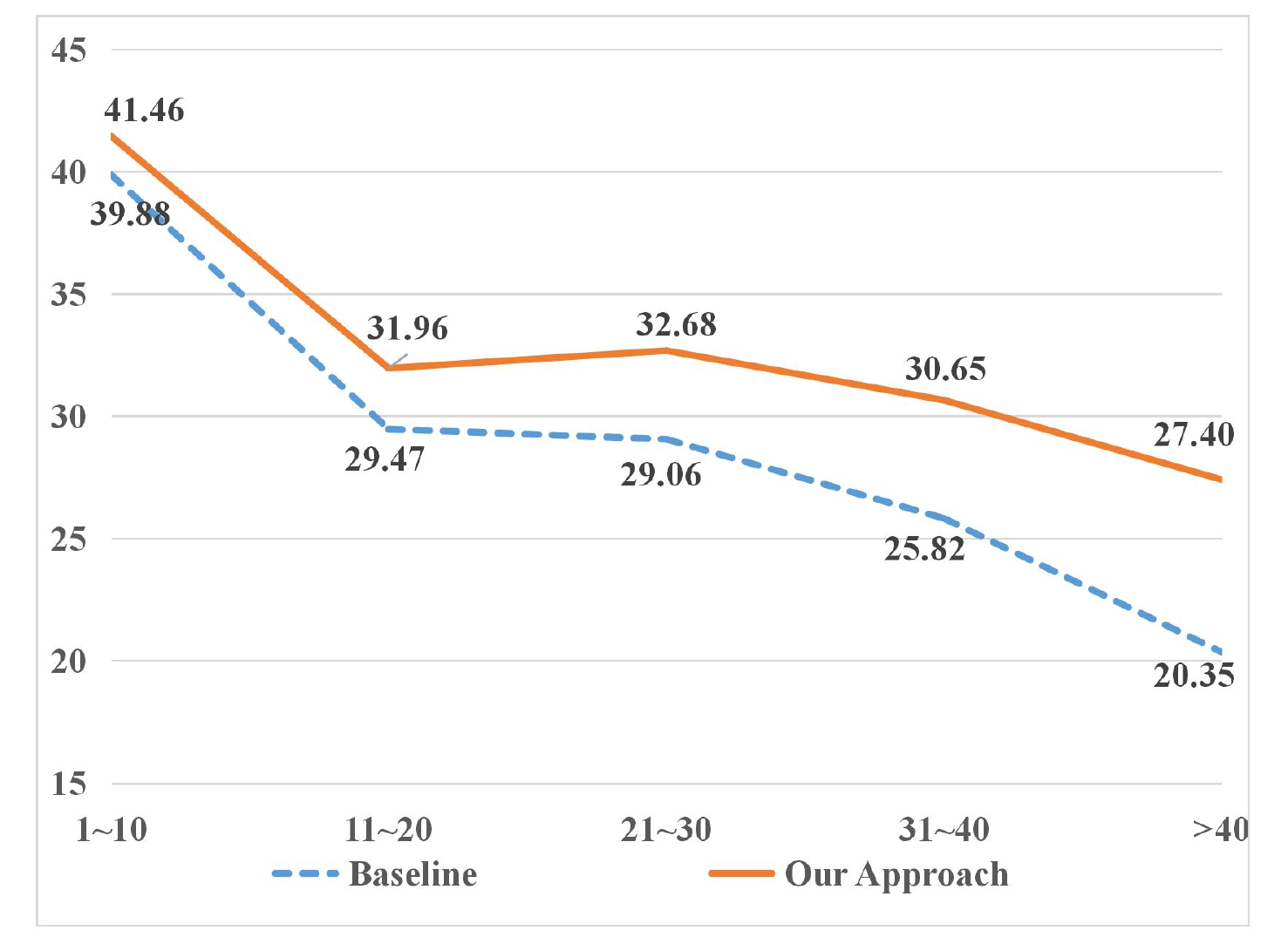}
\end{center}
\caption{Performance (in BLEU) on the test set with respect to the size of the input AMR graphs.} \label{fig:length}
\end{figure}

When we encode an AMR graph with plenty concepts, linearizing it into a sequence tends to lose great amount of structural information. In order to test the hypothesis that graphs with more concepts contribute more to the improvement, we partition source AMR graphs to different groups by their sizes (i.e., numbers of concepts) and evaluate their performance respectively. Figure~\ref{fig:length} shows the results which indicate that modeling graph structures significantly outperforms the baseline over all AMR lengths. We also observe that the performance gap between the baseline and our approach increases when AMR graphs become big, revealing that the baseline seq2seq model is far from capturing deep structural details of big AMR graphs. Figure~\ref{fig:length} also indicates that text generation becomes difficult for big AMR graphs. We think that the low performance on big AMR graphs is mainly attributed to two reasons: 

\begin{itemize}
    \item Big AMR graphs are usually mapped into long sentences while seq2seq model tends to stop early for long inputs. As a result, the length ratio\footnote{Length ratio is the length of generation output, divided by the length of reference.} for AMRs with more than 40 concepts is 0.906, much lower than that for AMRs with less concepts. 
    \item Big AMR graphs are more likely to have reentrancies, which makes the generation more challenging. 
\end{itemize}

\subsection{Case Study}

\begin{figure*}[h]
\begin{center}
\includegraphics[width=5.0in]{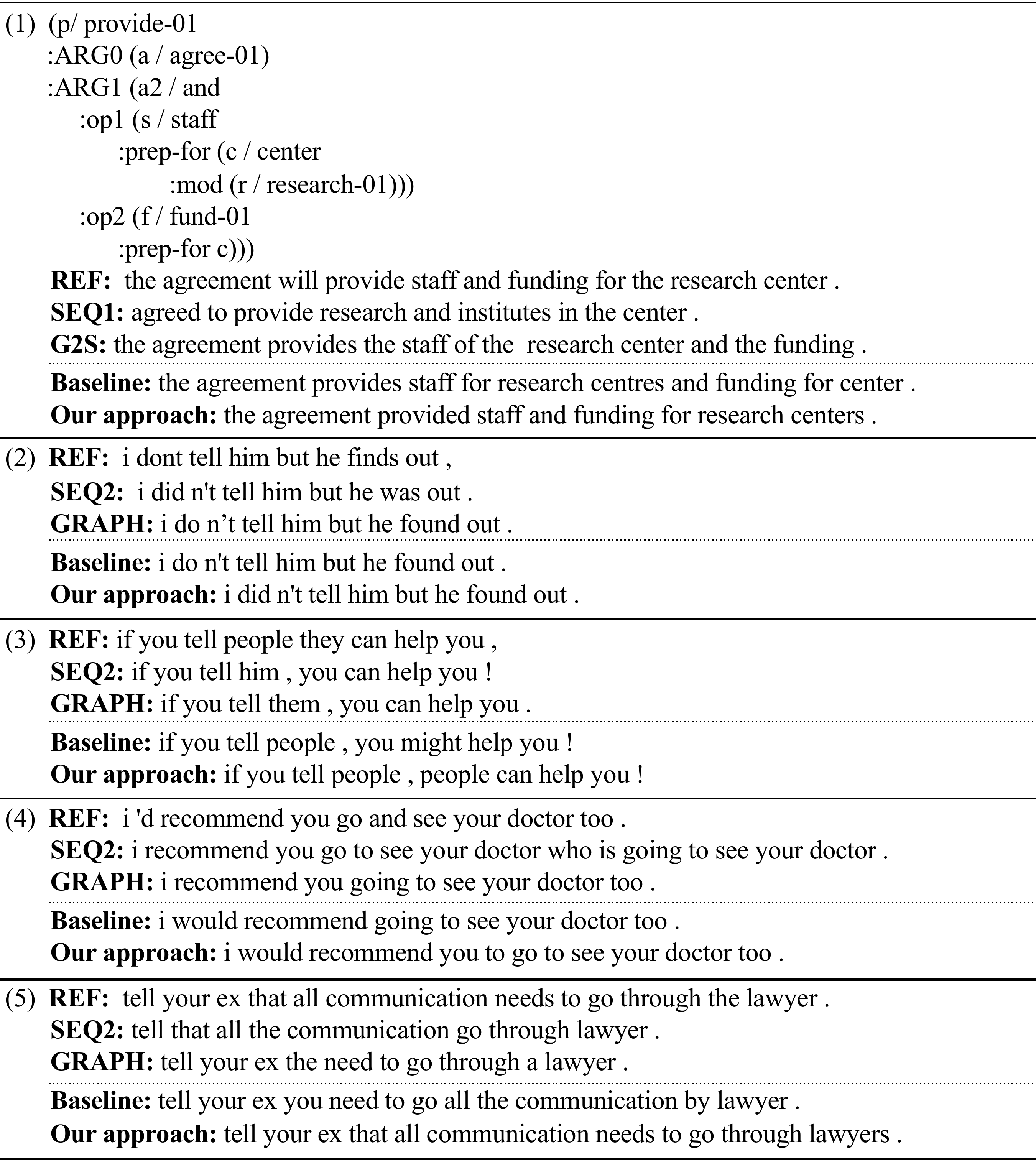}
\end{center}
\caption{Examples of generation from AMR graphs. (1) is from \citet{song_etal_acl:18}, (2) - (5) are from \citet{damonte_etal_naacl:19}. REF is the reference sentence. SEQ1 and G2S are the outputs of the seq2seq and the graph2seq models in \citet{song_etal_acl:18}, respectively. SEQ2 and GRAPH are the outputs of the seq2seq and the graph models in \citet{damonte_etal_naacl:19}, respectively. \ignore{Baseline and Structural Encoder are the output of our baseline and structural transformer, respectively.}} \label{fig:case}
\end{figure*}

In order to better understand the model performance, Figure~\ref{fig:case} presents a few examples studied in \citet{song_etal_acl:18} (Example (1)) and \citet{damonte_etal_naacl:19} (Examples (2) - (5)). 

In Example (1), though our baseline recovers a propositional phrase for the noun \textit{staff} and another one for the noun \textit{funding}, it fails to recognize the anaphora and antecedent relation between the two propositional phrases. In contrast, our approach successfully recognizes \textit{:prep-for c} as a reentrancy node and generates one propositional phrase shared by both nouns \textit{staff} and \textit{funding}. In Example (2), we note that although AMR graphs lack tense information, the baseline generates output with inconsistent tense (i.e., \textit{do} and \textit{found}) while our approach consistently prefers past tense for the two clauses. In Example (3), only our approach correctly uses \textit{people} as the subject of the predicate \textit{can}. In Example (4), the baseline fails to predict the direct object \textit{you} for predicate \textit{recommend}. Finally in Example (5), the baseline fails to recognize the subject-predicate relation between noun \textit{communicate} and verb \textit{need}. Overall, we note that compared to the baseline, our approach produces more accurate output and deal with reentrancies more properly.

Comparing the generation of our approach and graph-based models in \citet{song_etal_acl:18} and \citet{damonte_etal_naacl:19}, we observe that our generation is more close to the reference in sentence structure. Due to the absence of tense information in AMR graphs, our model tends to use past tense, as \textit{provided} and \textit{did} in Example (1) and (2). Similarly, without information concerning singular form and plural form, our model is more likely to use plural nouns, as \textit{centers} and \textit{lawyers} in Example (1) and (5).

\section{Related Work}

Most studies in AMR-to-text generation regard it as a translation problem and are motivated by the recent advances in both statistical machine translation (SMT) and neural machine translation (NMT). \citet{flanigan_etal_naacl:16} first transform an AMR graph into a tree, then specify a number of tree-to-string transduction rules based on alignments that are used to drive a tree-based SMT model~\cite{graehl_knight_naacl:04}. \citet{pourdamghani_etal:16} develop a method that learns to linearize AMR graphs into AMR strings, and then feed them into a phrase-based SMT model~\cite{koehn_etal_naacl:03}. \ignore{\citet{song_etal_emnlp:16}  recast generation as a traveling salesman problem, after partitioning the graph into fragments and finding the best linearization order. }\citet{song_etal_acl:17} use synchronous
node replacement grammar (SNRG) to generate text. Different from synchronous context-free grammar in hierarchical phrase-based SMT~\cite{chiang_cl:07}, SNRG is a grammar over graphs. 

Moving to neural seq2seq approaches, \citet{konstas_etal_acl:17} successfully apply seq2seq model together with large-scale unlabeled data for both text-to-AMR parsing and AMR-to-text generation. With special interest in the target side syntax, \citet{cao_etal_naacl:19} use seq2seq models to generate target syntactic structure, and then the surface form. To prevent the information loss in linearizing AMR graphs into sequences, \cite{song_etal_acl:18,beck_etal_acl:18} propose graph-to-sequence models to encode graph structure directly. Focusing on reentrancies, \citet{damonte_etal_naacl:19} propose stacking encoders which consist of BiLSTM~\cite{Graves_etal:13}, TreeLSTMs~\cite{tai_etal_acl:15}, and Graph Convolutional Network (GCN)~\cite{duvenaud_etal:15,kipf_and_welling:16}. \citet{guo_etal_tacl:19} propose densely connected GCN to better capture both local and non-local features. However, all the aforementioned graph-based models only consider the relations between nodes that are directly connected, thus lose the structural information between nodes that are indirectly connected via an edge path. 

Recent studies also extend the Transformer to encode structural information for other NLP applications. \citet{shaw_etal_naacl:18} propose relation-aware self-attention to capture relative positions of word pairs for neural machine translation. \citet{ge_etal_ijcai:19} extend the relation-aware self-attention to capture syntactic and semantic structures. Our model is inspired by theirs but aims to encode structural label sequences of concept pairs. \citet{kedziorski_etal_naacl:19} propose graph Transformer to encode graph structure. Similar to the GCN, it focuses on the relations between directly connected nodes. 

\section{Conclusion and Future Work}
In this paper we proposed a structure-aware self-attention for the task of AMR-to-text generation. The major idea of our approach is to encode long-distance relations between concepts in AMR graphs into the  self-attention encoder in the Transformer. In the setting of supervised learning, our models achieved the best experimental results ever reported on two English benchmarks. 

Previous studies have shown the effectiveness of using large-scale unlabelled data. In future work, we would like to do semi-supervised learning and use silver data to test how much improvements could be further achieved. 

\section*{Acknowledgments}
We thank the anonymous reviewers for their insightful comments and suggestions. We are grateful to Linfeng Song for fruitful discussions. This work is supported by the National Natural Science Foundation of China (Grant No. 61876120, 61673290, 61525205), and the
Priority Academic Program Development of Jiangsu Higher
Education Institutions.

\bibliographystyle{acl_natbib}
\bibliography{amr.bib}

\begin{thebibliography}{33}
\expandafter\ifx\csname natexlab\endcsname\relax\def\natexlab#1{#1}\fi

\bibitem[{Banarescu et~al.(2013)Banarescu, Bonial, Cai, Georgescu, Griffitt,
  Hermjakob, Knight, Koehn, Palmer, and Schneider}]{banarescu_etal:13}
Laura Banarescu, Claire Bonial, Shu Cai, Madalina Georgescu, Kira Griffitt, Ulf
  Hermjakob, Kevin Knight, Philipp Koehn, Martha Palmer, and Nathan Schneider.
  2013.
\newblock Abstract meaning representation for sembanking.
\newblock In \emph{Proceedings of 7th Linguistic Annotation Workshop \&
  Interoperability with Discourse}, pages 178--186.

\bibitem[{Banerjee and Lavie(2005)}]{banerjee_and_lavie:05}
Satanjeev Banerjee and Alon Lavie. 2005.
\newblock Meteor: An automatic metric for mt evaluation with improved
  correlation with human judgments.
\newblock In \emph{Proceedings of ACL}, pages 65--72.

\bibitem[{Beck et~al.(2018)Beck, Haffari, and Cohn}]{beck_etal_acl:18}
Daniel Beck, Gholamreza Haffari, and Trevor Cohn. 2018.
\newblock Graph-to-sequence learning using gated graph neural networks.
\newblock In \emph{Proceedings of ACL}, pages 273--283.

\bibitem[{Cao and Clark(2019)}]{cao_etal_naacl:19}
Kris Cao and Stephen Clark. 2019.
\newblock Factorising amr generation through syntax.
\newblock In \emph{Proceedings of NAACL}, pages 2157--2163.

\bibitem[{Chiang(2007)}]{chiang_cl:07}
David Chiang. 2007.
\newblock Hierarchical phrase-based translation.
\newblock \emph{Computational Linguistics}, 33:201--228.

\bibitem[{Damonte and Cohen(2019)}]{damonte_etal_naacl:19}
Marco Damonte and Shay~B. Cohen. 2019.
\newblock Structural neural encoders for {AMR}-to-text generation.
\newblock In \emph{Proceedings of NAACL}, pages 3649--3658.

\bibitem[{Denkowski and Lavie(2014)}]{denkowski_and_lavie:14}
Michael Denkowski and Alon Lavie. 2014.
\newblock Meteor universal: Language specific translation evaluation for any
  target language.
\newblock In \emph{Proceedings of WMT}, pages 376--380.

\bibitem[{Duvenaud et~al.(2015)Duvenaud, Maclaurin, Iparraguirre, Bombarell,
  Hirzel, Aspuru-Guzik, and Adams}]{duvenaud_etal:15}
David~K Duvenaud, Dougal Maclaurin, Jorge Iparraguirre, Rafael Bombarell,
  Timonthy Hirzel, Alan Aspuru-Guzik, and Ryan~P Adams. 2015.
\newblock Convolutional networks on graphs for learning molecular fingerprints.
\newblock In \emph{Proceedings of NIPS}, pages 2224--2232.

\bibitem[{Flanigan et~al.(2016)Flanigan, Dyer, Smith, and
  Carbonell}]{flanigan_etal_naacl:16}
Jeffrey Flanigan, Chris Dyer, Noah~A. Smith, and Jaime Carbonell. 2016.
\newblock Generation from abstract meaning representation using tree
  transducers.
\newblock In \emph{Proceedings of NAACL}, pages 731--739.

\bibitem[{Ge et~al.(2019)Ge, Li, Zhu, and Li}]{ge_etal_ijcai:19}
DongLai Ge, Junhui Li, Muhua Zhu, and Shoushan Li. 2019.
\newblock Modeling source syntax and semantics for neural amr parsing.
\newblock In \emph{Proceedings of IJCAI}, pages 4975--4981.

\bibitem[{Graehl and Knight(2004)}]{graehl_knight_naacl:04}
Jonathan Graehl and Kevin Knight. 2004.
\newblock Training tree transducers.
\newblock In \emph{Proceedings of NAACL}, pages 105--112.

\bibitem[{Graves et~al.(2013)Graves, rahman Mohamed, and
  Hinton}]{Graves_etal:13}
Alex Graves, Abdel rahman Mohamed, and Geoffrey Hinton. 2013.
\newblock Speech recognition with deep recurrent neural networks.
\newblock In \emph{Proceedings of ICASSP}, pages 6645--6649.

\bibitem[{Gulcehre et~al.(2016)Gulcehre, Ahn, Nallapati, Zhou, and
  Bengio}]{gulcehre_etal_acl:16}
Caglar Gulcehre, Sungjin Ahn, Ramesh Nallapati, Bowen Zhou, and Yoshua Bengio.
  2016.
\newblock Pointing the unknown words.
\newblock In \emph{Proceedings of ACL}, pages 140--149.

\bibitem[{Guo et~al.(2019)Guo, Zhang, Teng, and Lu}]{guo_etal_tacl:19}
Zhijiang Guo, Yan Zhang, Zhiyang Teng, and Wei Lu. 2019.
\newblock Densely connected graph convolutional networks for graph-to-sequence
  learning.
\newblock \emph{Transactions of the Association of Computational Linguistics},
  7:297--312.

\bibitem[{Kalchbrenner et~al.(2014)Kalchbrenner, Grefenstette, and
  Blunsom}]{kalchbrenner_etal_acl:14}
Nal Kalchbrenner, Edward Grefenstette, and Phil Blunsom. 2014.
\newblock A convolutional neural network for modelling sentences.
\newblock In \emph{Proceedings of ACL}, pages 655--665.

\bibitem[{Kingma and Ba(2015)}]{kingma_ba_iclr:15}
Diederik~P. Kingma and Jimmy Ba. 2015.
\newblock Adam: A method for stochastic optimization.
\newblock In \emph{Proceedings of ICLR}.

\bibitem[{Kipf and Welling(2016)}]{kipf_and_welling:16}
Thomas~N Kipf and Max Welling. 2016.
\newblock Semi-supervised classification with graph convolutional networks.
\newblock In \emph{Proceedings of ICLR}.

\bibitem[{Klein et~al.(2017)Klein, Kim, Deng, Senellart, and
  Rush}]{klein_etal:17}
Guillaume Klein, Yoon Kim, Yuntian Deng, Jean Senellart, and Alexander~M. Rush.
  2017.
\newblock Opennmt: Open-source toolkit for neural machine translation.
\newblock In \emph{Proceedings of ACL, System Demonstrations}, pages 67--72.

\bibitem[{Koehn(2004)}]{koehn_emnlp:04}
Philipp Koehn. 2004.
\newblock Statistical significance tests for machine translation evaluation.
\newblock In \emph{Proceedings of EMNLP}, pages 388–--395.

\bibitem[{Koehn et~al.(2003)Koehn, Och, and Marcu}]{koehn_etal_naacl:03}
Philipp Koehn, Franz~J. Och, and Daniel Marcu. 2003.
\newblock Statistical phrase-based translation.
\newblock In \emph{Proceedings of NAACL}, pages 127--133.

\bibitem[{Koncel-Kedziorski et~al.(2019)Koncel-Kedziorski, Bekal, Luan, Lapata,
  and Hajishirzi}]{kedziorski_etal_naacl:19}
Rik Koncel-Kedziorski, Dhanush Bekal, Yi~Luan, Mirella Lapata, and Hannaneh
  Hajishirzi. 2019.
\newblock Text generation from knowledge graphs with graph transformers.
\newblock In \emph{Proceedings of NAACL}, pages 2284--2293.

\bibitem[{Konstas et~al.(2017)Konstas, Iyer, Yatskar, Choi, and
  Zettlemoyer}]{konstas_etal_acl:17}
Ioannis Konstas, Srinivasan Iyer, Mark Yatskar, Yejin Choi, and Luke
  Zettlemoyer. 2017.
\newblock Neural {AMR}: Sequence-to-sequence models for parsing and generation.
\newblock In \emph{Proceedings of ACL}, pages 146--157.

\bibitem[{Lin et~al.(2017)Lin, Feng, dos Santos, Yu, Xiang, Zhou, and
  Bengio}]{lin_etal_iclr:17}
Zhouhan Lin, Minwei Feng, Cicero~Nogueira dos Santos, Mo~Yu, Bing Xiang, Bowen
  Zhou, and Yoshua Bengio. 2017.
\newblock A structured self-attentive sentence embedding.
\newblock In \emph{Proceedings of ICLR}.

\bibitem[{Papineni et~al.(2002)Papineni, Roukos, Todd, and
  Zhu}]{papineni_etal_acl:02}
Kishore Papineni, Salim Roukos, Ward Todd, and Wei-Jing Zhu. 2002.
\newblock Bleu: a method for automatic evaluation of machine translation.
\newblock In \emph{Proceedings of ACL}, pages 311--318.

\bibitem[{Popović(2017)}]{popovic_wmt:17}
Maja Popović. 2017.
\newblock chrf++: words helping character n-grams.
\newblock In \emph{Proceedings of WMT}, pages 612--618.

\bibitem[{Pourdamghani et~al.(2016)Pourdamghani, Knight, and
  Hermjakob}]{pourdamghani_etal:16}
Nima Pourdamghani, Kevin Knight, and Ulf Hermjakob. 2016.
\newblock Generating english from abstract meaning representations.
\newblock In \emph{Proceedings of the 9th International Natural Language
  Generation conference}, pages 21--25.

\bibitem[{Sennrich et~al.(2016)Sennrich, Haddow, and
  Birch}]{sennrich_etal_acl:16}
Rico Sennrich, Barry Haddow, and Alexandra Birch. 2016.
\newblock Neural machine translation of rare words with subword units.
\newblock In \emph{Proceedings of ACL}, pages 1715--1725.

\bibitem[{Shaw et~al.(2018)Shaw, Uszkoreit, and Vaswani}]{shaw_etal_naacl:18}
Peter Shaw, Jakob Uszkoreit, and Ashish Vaswani. 2018.
\newblock Self-attention with relative position representations.
\newblock In \emph{Proceedings of NAACL}, pages 464–--468.

\bibitem[{Song et~al.(2017)Song, Peng, Zhang, Wang, and
  Gildea}]{song_etal_acl:17}
Linfeng Song, Xiaochang Peng, Yue Zhang, Zhiguo Wang, and Daniel Gildea. 2017.
\newblock Amr-to-text generation with synchronous node replacement grammar.
\newblock In \emph{Proceedings of ACL}, pages 7--13.

\bibitem[{Song et~al.(2016)Song, Zhang, Peng, Wang, and
  Gildea}]{song_etal_emnlp:16}
Linfeng Song, Yue Zhang, Xiaochang Peng, Zhiguo Wang, and Daniel Gildea. 2016.
\newblock Amr-to-text generation as a traveling salesman problem.
\newblock In \emph{Proceedings of EMNLP}, pages 2084--2089.

\bibitem[{Song et~al.(2018)Song, Zhang, Wang, and Gildea}]{song_etal_acl:18}
Linfeng Song, Yue Zhang, Zhiguo Wang, and Daniel Gildea. 2018.
\newblock A graph-to-sequence model for {AMR}-to-text generation.
\newblock In \emph{Proceedings of ACL}, pages 1616--1626.

\bibitem[{Tai et~al.(2015)Tai, Socher, and Manning}]{tai_etal_acl:15}
Kai~Sheng Tai, Richard Socher, and Christopher~D. Manning. 2015.
\newblock Improved semantic representations from tree-structured long
  short-term memory networks.
\newblock In \emph{Proceedings of ACL}, pages 1556--1566.

\bibitem[{Vaswani et~al.(2017)Vaswani, Shazeer, Parmar, Uszkoreit, Jones,
  N.Gomez, Kaiser, and Polosukhin}]{vaswani_etal_nips:17}
Ashish Vaswani, Noam Shazeer, Niki Parmar, Jakob Uszkoreit, Llion Jones, Aidan
  N.Gomez, Lukasz Kaiser, and Illia Polosukhin. 2017.
\newblock Attention is all you need.
\newblock In \emph{Proceedings of NIPS}, pages 5998--6008.

\end{thebibliography}

\end{document}